\documentclass[10pt,twocolumn,letterpaper]{article}

\usepackage[pagenumbers]{cvpr} 

\usepackage{graphicx}
\usepackage{amsmath}
\usepackage{amssymb}
\usepackage{booktabs}
\usepackage{multirow}
\usepackage{enumitem}

\usepackage[pagebackref,breaklinks,colorlinks]{hyperref}

\usepackage[capitalize]{cleveref}
\crefname{section}{Sec.}{Secs.}
\Crefname{section}{Section}{Sections}
\Crefname{table}{Table}{Tables}
\crefname{table}{Tab.}{Tabs.}

\def\cvprPaperID{***} 
\def\confName{CVPR}
\def\confYear{2022}

\begin{document}
	\hyphenpenalty=4000
	\tolerance=1000
	
	\title{HP-Capsule: Unsupervised Face Part Discovery by Hierarchical Parsing \\ Capsule Network}
	
	\author{Chang Yu$^{1,2}$, Xiangyu Zhu$^{1,2}$, Xiaomei Zhang$^{1,2}$, Zidu Wang$^{1,2}$, Zhaoxiang Zhang$^{1,2,3}$, Zhen Lei$^{1,2,3}$\footnotemark[1]\\
		$^{1}$NLPR, Institute of Automation, Chinese Academy of Sciences\\
		$^{2}$School of Artificial Intelligence, University of Chinese Academy of Sciences\\
		$^{3}$ Centre for Artificial Intelligence and Robotics, Hong Kong Institute of Science \& Innovation,\\ Chinese Academy of Sciences\\
		{\tt\small \{chang.yu, xiangyu.zhu, zlei\}@nlpr.ia.ac.cn}\\
		{\tt\small \{zhangxiaomei2016, wangzidu2022, zhaoxiang.zhang\}@ia.ac.cn}\\
	} 

	\twocolumn[{%
		\renewcommand\twocolumn[1][]{#1}%
		\maketitle
		\centering
		\includegraphics[width=0.9\linewidth]{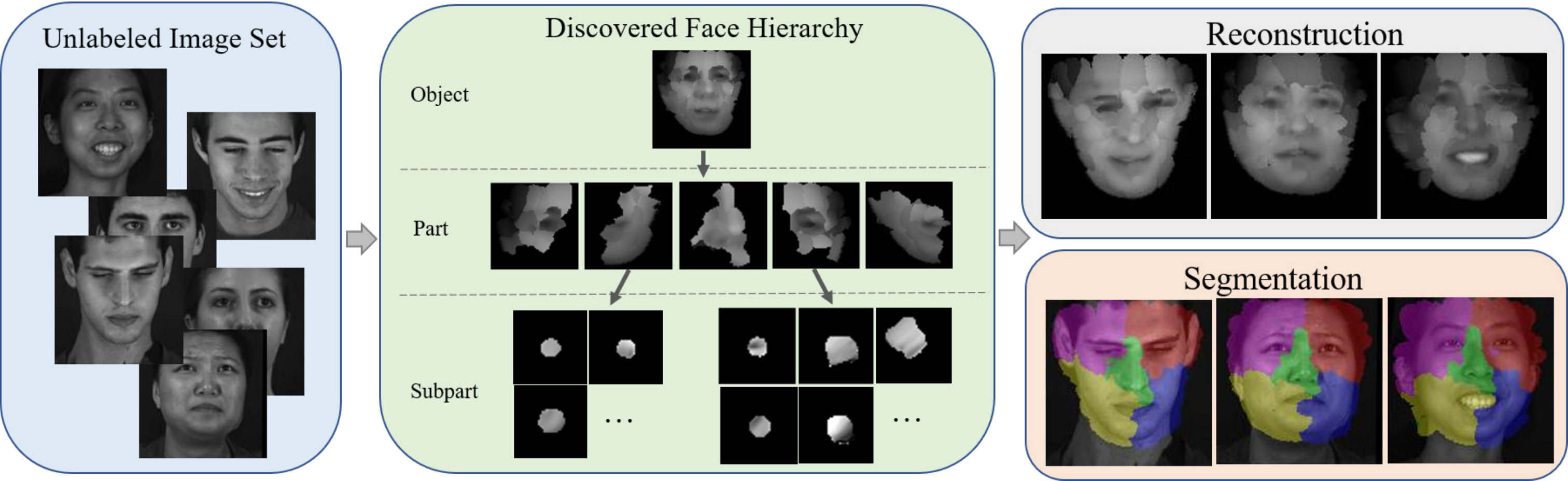}		
		\captionof{figure}{\label{fig:fig1}
			A brief review of the Hierarchical Parsing Capsule Network (HP-Capsule). Given a large scale of unlabeled images (left), HP-Capsule can automatically discover the hierarchical face parts (middle) and give unsupervised face segmentation results as by-products (right).}
		\vspace{20px}
	}]
	
	{\renewcommand{\thefootnote}{\fnsymbol{footnote}}
		\footnotetext[1]{Corresponding author.}}
	\begin{abstract}
		Capsule networks are designed to present the objects by a set of parts and their relationships, which provide an insight into the procedure of visual perception. Although recent works have shown the success of capsule networks on simple objects like digits, the human faces with homologous structures, which are suitable for capsules to describe, have not been explored. In this paper, we propose a Hierarchical Parsing Capsule Network (HP-Capsule) for unsupervised face subpart-part discovery. When browsing large-scale face images without labels, the network first encodes the frequently observed patterns with a set of explainable subpart capsules. Then, the subpart capsules are assembled into part-level capsules through a Transformer-based Parsing Module (TPM) to learn the compositional relations between them. During training, as the face hierarchy is progressively built and refined, the part capsules adaptively encode the face parts with semantic consistency. HP-Capsule extends the application of capsule networks from digits to human faces and takes a step forward to show how the neural networks understand homologous objects without human intervention. Besides, HP-Capsule gives unsupervised face segmentation results by the covered regions of part capsules, enabling qualitative and quantitative evaluation. Experiments on BP4D and Multi-PIE datasets show the effectiveness of our method.

	\end{abstract}
	
	\section{Introduction}
	\label{sec:intro}
	Psychological studies\cite{marr1978representation,hinton1979some,singh2001part} reveal that: The recognition procedure is often assigned with hierarchical structural descriptions by parsing the shapes into components and organizing them with their spatial relationships. This statement is consistent with the improvements of various perceptual tasks which incorporate part-level information for better embeddings\cite{chen2018knowledge,quan2019auto,xie2020adversarial}. However, most of the existing methods implement the parsing in a predefined way~\cite{zhang2020part,zhao2018psanet,lin2019face,liu2020new}, where the definitions of parts are given by humans. Such handcrafted parsing can not reflect how neural networks understand objects.
	
	To explore the visual perception mechanism of neural networks, an intriguing way is to learn the parsing directly from the data. The network should discover the visual part concepts with as little human intervention as possible and keep the semantic consistency across different samples, as shown in Figure~\ref{fig:fig1}. This unsupervised learning task is still a challenging problem, as the semantic parts are difficult to be described mathematically.
	
\begin{figure}
		\centering
		\includegraphics[width=0.95\linewidth]{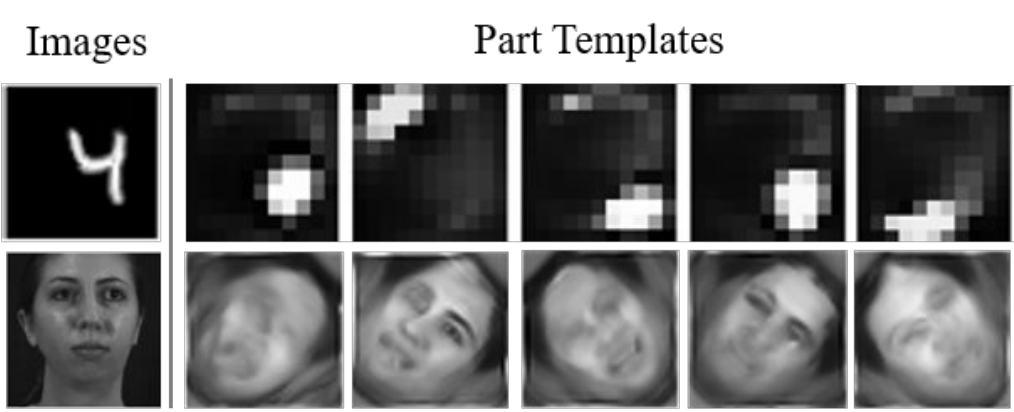}	
		\caption{The parts discovered by SCAE~\cite{kosiorek2019stacked} on digits and faces. Even though SCAE can capture parts on simple digits, it tends to fail on more complicated objects like human faces, learning holistic templates as parts.}
		\label{fig-scae}
	\end{figure}

	Capsule networks, which are designed to present objects by a set of parts and their relationships, are a feasible solution for this unsupervised parsing task. Among the capsule structures proposed in recent years~\cite{hinton2011transforming,sabour2017dynamic,hinton2018matrix,kosiorek2019stacked,sabour2021unsupervised}, SCAE~\cite{kosiorek2019stacked} is most suitable for this unsupervised face part discovery task since it presents objects with a part-whole hierarchy and defines the capsule as a set of explainable parameters including presences, poses, and visualizable templates. However, SCAE can only parse simple objects like handwritten digits. When extended to face images, SCAE fails to capture face parts and decomposes faces as a whole, generating holistic representations in the part capsules, as shown in Figure~\ref{fig-scae}. 
	
	In this paper, we will improve the capsule network to handle human faces, which have homologous structures but diverse appearances. A Hierarchical Parsing Capsule Network (HP-Capsule) is proposed to discover hierarchical face parts and their relationships directly from the unlabeled image sets. Specifically, HP-Capsule understands faces with two sub-modules: a capsule-based autoencoder for subpart discovery and a Transformer-based Parsing Module (TPM) for hierarchy construction. During training, the frequently observed patterns are first captured and encoded by the capsule encoder with a set of explainable capsule parameters. A Visibility Activation Function (VAF) is proposed to constrain the process of reconstructing the input image with capsule templates, so that the object is spatially decomposed into subparts rather than holistic representations. Then, the discovered subparts are regarded as visual words and sent to TPM to be aggregated to higher-level part capsules, where the subpart-to-part hierarchy is naturally built. Several constraints are incorporated to preserve the shape and appearance consistency so that the generated parts have more prominent semantics. 
	
	As a by-product, the covered regions of part capsules can be regarded as the segmentation maps, which can be used for the unsupervised face segmentation task, enabling the evaluation of our method. Compared with other unsupervised segmentation methods, HP-Capsule shows better semantic consistency and provides more interpretable descriptions about the discovered parts, including the visualizable templates and the statistics on presence and pose. 
	
	To summarize, the main contributions of this work are:
	
	\begin{itemize}
		\item This paper proposes a Hierarchical Parsing Capsule Network (HP-Capsule) for unsupervised face hierarchy discovery. HP-Capsule provides an insight into how the neural network understands homologous structures without human interventions. 
		
		\item In the subpart discovery process, we propose a Visibility Activation Function (VAF), which enforces the network to concentrate on template regions with higher visibility, to ensure the objects are decomposed into localized subparts rather than holistic representations.
		
		\item A Transformer-based Parsing Module (TPM) is proposed to aggregate subparts into parts, constructing the subpart-part face hierarchy. The covered regions of part capsules can also be used for unsupervised face segmentation. Experiments on BP4D and Multi-PIE show the effectiveness of our method.
		
	\end{itemize}

	\medskip

	\begin{figure*}
		\centering
		\includegraphics[width=0.95\linewidth]{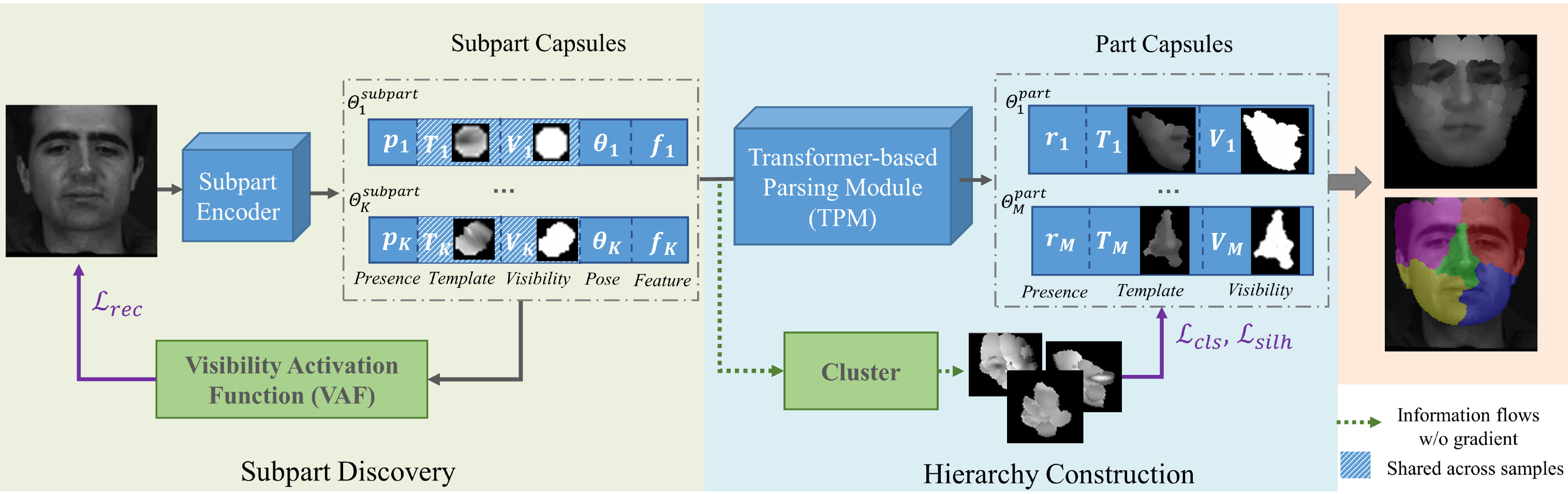}
		
		\caption{Overview of the Hierarchical Parsing Capsule Network (HP-Capsule). HP-Capsule understands faces with two sub-modules: the capsule autoencoder with Visibility Activation Function (VAF) for subpart discovery and the Transformer-based Parsing Module (TPM) for hierarchy construction. During training, the frequently observed patterns are firstly captured and encoded with a set of explainable capsule parameters, which are then passed through the VAF for subpart discovery. After that, the explored subpart capsules are sent into the TPM for higher-level part capsules. TPM is trained with the shape consistency loss and the pseudo subpart-part relationships generated by clustering. By automatically aggregating subparts into parts, the face hierarchy is naturally built.}
		\label{fig:overview}
	\end{figure*}

	\section{Related Work}
	\label{sec:related}
	
	\textbf{Capsule Networks.} Inspired by the sparse connection of human brains, 
	capsule networks are designed to present the objects with a dynamic parse tree. Given an input image, the capsule network would automatically activate some of the capsules and then pass them to the next layers. When assigning the capsules with explicit meanings, the interpretability of the network would be improved. Recent works have shown the success of capsules on various tasks~\cite{zhao20193d,xin2020facial,singh2019dual,wu2019speech} but most of them use capsules as the enhanced MLPs, while the interpretability, as well as the parsing characteristics of capsules, have not been well explored. For years, capsule networks have evolved many versions~\cite{hinton2011transforming,sabour2017dynamic,hinton2018matrix,kosiorek2019stacked,sabour2021unsupervised}. Among them, SCAE~\cite{kosiorek2019stacked} is an intriguing structure that describes the objects with a set of visualizable templates through unsupervised learning. However, SCAE can only tackle simple objects like handwritten digits where the learned templates are stroke-like. Recently, Sabour \etal.~\cite{sabour2021unsupervised} extend the capsule network for complicated images like human bodies, but it needs optical flow as clues and can only show the mask of each part. In this paper, we extend SCAE to more complicated objects like human faces, discovering the face hierarchy directly from the unlabeled image sets.

	\textbf{Unsupervised Part Discovery.} Unsupervised learning has witnessed impressive progress in recent years~\cite{wu2020unsupervised,wu2019unsupervised,wu2020end,he2020momentum,chen2020simple,van2021unsupervised,cho2021picie,weng2021unsupervised}. Our work is related to the task of unsupervised part discovery, which aims to discover visual concepts from unlabeled images or videos. In the early years, Feng \etal.~\cite{feng2002local} propose the local non-negative matrix factorization (LNMF) to learn part-based representations for human faces. By adding localized sparsity factorization, LNMF can learn part-based basis, but the performance is still constrained as the faces are presented in the linear space. Recent works~\cite{bau2017network,gonzalez2018semantic} show that the semantic parts have already been included in the CNN features. Inspired by this, Collins \etal.~\cite{collins2018deep} propose to use non-negative matrix factorization (NMF) on CNN activations to locate the semantic concepts on image sets. However, this approach needs to solve optimization during inference and lacks interpretability. Besides using activation maps, other methods~\cite{Xu2018modeling,sabour2021unsupervised,Gao2021ucos} try to use motion clues in videos to discover parts as the regions with the same semantics always move together. Differently, our method does not need motion clues and the learned parts can be visualized directly to gain better explainability.
	
	\textbf{Unsupervised Face Segmentation.} Recently, several works have made positive progress on the unsupervised face segmentation task. Hung \etal.~\cite{hung2019scops} propose several loss functions to learn the face parts that are geometrically concentrated and robust to spatial transformations. Their method needs saliency maps to suppress the background features and Liu \etal.~\cite{liu2021unsupervised} argue that this might make the method less reliable. As a result, they propose a concentration loss to separate the background and a squeeze-and-expand block for better shape representations. Gao \etal.~\cite{Gao2021ucos} propose a dual procedure by leveraging the motion clues embedded in videos to further improve the performance. Compared with these methods, our networks are more explainable and can obtain parts with better semantic consistency.
	
	\section{Methodology}
	\label{sec:formatting}
	Given a collection of images from the same category, our work aims to learn a model that can discover hierarchical parts and their relationships from a single image. In the following parts, we will introduce the overall framework in Section~\ref{subsec:framwork}, the capsule autoencoder with Visibility Activation Function for subpart discovery in Section~\ref{subsec:AAActive}, and the Transformer-based Parsing Module for hierarchy construction in Section~\ref{subsec:transformer}.
	
	\subsection{Overall Framework}
	\label{subsec:framwork}
	In this paper, we propose a Hierarchical Parsing Capsule Network for unsupervised face hierarchy discovery, which includes a capsule autoencoder for subpart discovery and a parsing module for hierarchy construction. The overall framework is shown in Figure~\ref{fig:overview}.
	
	Given an input image $\mathbf{I}$, the network first uses a capsule encoder to estimate $K$ subpart capsules:
	\begin{equation}
		\label{eq-subpart-caps}
		\Theta^{s}_1,\Theta^{s}_2,..., \Theta^{s}_K = \mathcal{E}_{enc}(\mathbf{I}),
	\end{equation}
	where each subpart capsule $\Theta^{s}_k$ is a set of parameters with explainable physical meanings: $\Theta^{s}_k:\{p_k^s, \theta_k^s, T_k^s, V_k^s, f_k^s\}, k\in[1, K]$, including presence probability $p_k^s\in \mathbb{R}^{1\times1}$, pose $\theta_k^s\in \mathbb{R}^{1\times6}$, template $T_k^s\in \mathbb{R}^{H^s\times W^s}$, visibility map $V_k^s\in \mathbb{R}^{H^s\times W^s}$ and input-specific feature $f_k^s\in \mathbb{R}^{1\times D}$. Particularly, $T_k^s$ and $V_k^s$ describe the shape and the visible region of the discovered subparts, which are shared among samples.

	The discovered subpart capsules are regarded as visual words and are sent into a transformer-based parsing module to estimate higher-level part capsules, where the subpart-to-part hierarchy is naturally built:
	\begin{equation}
		\label{eq-part-caps}
		\Theta^{p}_1,\Theta^{p}_2,..., \Theta^{p}_M= \mathcal{E}_{TPM}(\Theta^{s}_1, \Theta^{s}_2, ..., \Theta^{s}_K),
	\end{equation}
	where the $m$-th part capsule $\Theta^{p}_m:\{T_m^p, V_m^p, r_m^p\}$ includes template $T_m^p\in \mathbb{R}^{H^p\times W^p}$, visibility map $V_m^p\in \mathbb{R}^{H^p\times W^p}$ and relationship $r_m^p\in \mathbb{R}^{1\times K}$ that describes the compositional relations between all subpart capsules. The $T_m^p$ and $V_m^p$ are aggregated from subparts according to the relationship $r_m^p$. When the face hierarchy is constructed, the part capsules naturally encode the discovered face parts with semantics, whose covered regions can also be used for unsupervised face segmentation.
	

	\subsection{Subpart Discovery by Visibility Activation Function}
	\label{subsec:AAActive}
	\begin{figure}
		\centering
		\begin{subfigure}{1.0\linewidth}
			\includegraphics[width=0.9\linewidth]{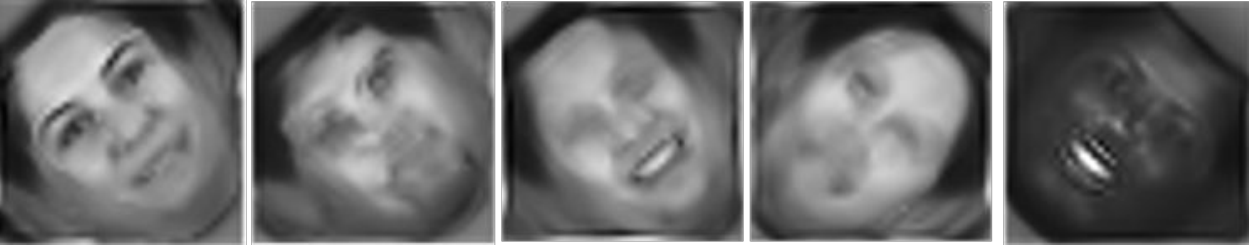}
			\caption{The subpart templates discovered w/o VAF. }
			\label{fig:template-a}
		\end{subfigure}
		\vspace{8px}
		
		\begin{subfigure}{1.0\linewidth}
			\includegraphics[width=0.9\linewidth]{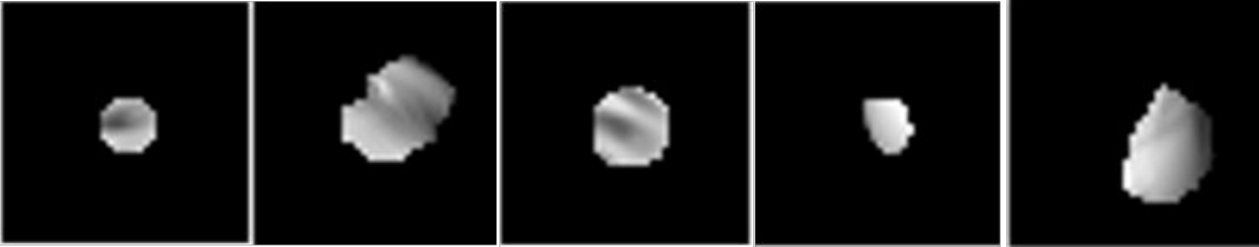}
			\caption{The subpart templates discovered with VAF. }
			\label{fig:template-b}
		\end{subfigure}
		\caption{The impact of the Visibility Activation Function (VAF). Compared with the holistic templates in (a), the subpart templates discovered in (b) are sparse and local-connected.}
		\label{fig:template}
	\end{figure}
	
	Parts can be roughly defined as sparse and local-connected regions that are semantically consistent across objects\cite{gonzalez2018semantic,hung2019scops}. The recent capsule network SCAE\cite{kosiorek2019stacked} has shown promising results on simple objects like digits, but tends to fail on more complicated objects like human faces. As shown in Figure~\ref{fig:template-a}, SCAE captures holistic face representations as the templates in subpart capsules. To enable the learning of part-level templates, we propose a Visibility Activation Function (VAF) for the capsule autoencoder.
	
	During training, the network firstly uses a self-attention based capsule encoder\cite{kosiorek2019stacked} to estimate the parameters of subpart capsules $\Theta^{s}_k:\{p_k^s, \theta_k^s, T_k^s, V_k^s, f_k^s\}$. Secondly, the color component $C_k^s$ is decoded from feature $f_k^s$ for template refinement. Thirdly, template $T_k^s$ and visibility map $V_k^s$ are transformed by the pose parameter $\theta^{s}_k$. Finally, the transformed $\hat{T}_k^s$ and $\hat{V}_k^s$ are passed through VAF to get the activated template $\tilde{T}_k^s$. This capsule encoding process can be formulated as:
	\begin{equation}\label{equ-subpart}
		\begin{aligned}
			p_k^s, \theta_k^s, T_k^s, V_k^s, f_k^s &= \mathcal{E}_{enc}(\mathbf{I}),\\
			C_s^k &= \mathrm{MLP}\left(f_k^s\right),\\
			\hat{T}_k^s, \hat{V}_k^s &= \mathrm{AffineTrans}(T_k^s, V_k^s, \theta_k^s),\\
			\tilde{T}_k^s &= \mathbf{VAF}\left(\hat{T}_k^s, \hat{V}_k^s\right).
		\end{aligned}
	\end{equation}
	
	Considering the visibility map $\hat{V}^s$ controls the silhouette of subparts, we employ VAF to suppress the learning of invisible regions, which is formulated as:
	\begin{align}\label{equ-VAF}
		\tilde{T}_{k_{i,j}}^s=
		\left\{
		\begin{aligned}
			&\hat{T}_{k_{i,j}}^s,  ~~~ if ~  \hat{V}_{k_{i,j}}^s \ge \gamma\\
			&-1,  ~~~~ if ~ \hat{V}_{k_{i,j}}^s < \gamma,
		\end{aligned}
		\right.
	\end{align}
	where $i,j$ stands for the position of the pixel and $\gamma$ is the hyper-parameter threshold for activation. 

	VAF encourages the network to reconstruct the input image only with the high-visibility template regions, so that the non-activated regions with low visibility, which are always represented as the blurred mean face, are suppressed without receiving back-propagated signals during training. 
	By Eqn.~\ref{equ-subpart}, the templates of subpart capsules can be transformed into the image space. If the input image $\mathbf{I}$ is modeled as a Gaussian Mixture of subpart capsules, the reconstruction loss is given by:
	\begin{equation}\label{eq-gauss-subpart}
		\mathcal{L}_{rec} = -\prod_{i,j} \sum_{k=1}^K  p_k^s \hat{V}_{k_{i,j}}^s \mathcal{N}(I_{i,j}\mid C_{k_{i,j}}^s\cdot\tilde{T}_{k_{i,j}}^s; \sigma_k^2),
	\end{equation}
	where $\sigma_k^2$ presents the variance of the Gaussian Mixture.
	
	We also incorporate $\mathcal{L}_{pres}$ to constrain the sparisty of the activated subpart capsules:
	\begin{small}
		\begin{equation}\label{equ-pres-s}
			\mathcal{L}_{pres} = \frac{1}{B}\sum_{b=1}^{B}\left(\sum_{k=1}^{K}p^s_{k,b}-\tau\right)^2+\frac{1}{K}\sum_{k=1}^{K}\left(\sum_{b=1}^{B}p^s_{k,b}-\frac{\tau B}{K}\right)^2,
		\end{equation} 
	\end{small}where $B$ stands for the batch size and $\tau$ is the hyper-parameter for the average number of activated subpart capsules. The network optimized by Eqn.~\ref{equ-pres-s} will try to activate the same number but different subpart capsules during the mini-batch. 
	
	Besides, we employ two losses to constrain the connectivity of the discovered subparts, including the $\mathcal{L}_{cen}$ for geometrical concentration~\cite{hung2019scops} and $\mathcal{L}_{std}$ for balancing the visible regions of different subpart templates:
	\begin{small}
		\begin{equation}\label{equ-Lvis}
			\begin{aligned}
				&\mathcal{L}_{cen} = \sum_{k}\sum_{i,j}\Vert (i,j)-(W^s/2, H^s/2)\Vert_2 \cdot V^s_{k_{i,j}}, \\
				&\mathcal{L}_{std} = std \left(\sum_{i,j}V^s_{k_{i,j}} \right),
			\end{aligned}
		\end{equation}
	\end{small}where $W^s$ and $H^s$ are the width and height of subpart templates, $std$ stands for the standard deviation.
	
	The loss function for subpart discovery is combined as:
	\begin{equation}
		\begin{aligned}\label{equ-subpart-loss}
			&\mathcal{L}_{subpart} = \lambda_{rec}\mathcal{L}_{rec} + \lambda_{pres}\mathcal{L}_{pres} +\lambda_{cen}\mathcal{L}_{cen} + \lambda_{std} \mathcal{L}_{std}.
		\end{aligned}
	\end{equation}
	
	As shown in Figure \ref{fig:template-b}, the network with VAF can effectively discover the localized appearance of faces.
	\subsection{Hierarchy Construction via Transformer-based Parsing Module}
	\label{subsec:transformer}
	The affine transformation and compositions of small subparts make them feasible enough to handle pose and appearance variations. However, they have ambiguous semantics due to their limited expressive power. As shown in Figure~\ref{fig-heatmap}, the discovered skin template may respond to both forehead and cheekbone and the eyes-related template highlights both left and right eyes. Therefore, if the subparts are assembled with appearance as clues only, there will be ambiguity in the face hierarchy, leading to flaws of the aggregated parts. 
	
	We propose a Transformer-based Parsing Module (TPM) to automatically construct the stable subpart-part hierarchy. 
	If we regard each subpart capsule as a visual word, the collection of subpart capsules becomes a sentence, which is very long since we need numbers of subparts (75 in this paper) to handle the complicated appearance of faces. 
	Tranformer~\cite{vaswani2017attention}, which is originally proposed to capture the long-range dependencies across words in natural language processing, is naturally extendable to our task. 
	In this paper, we adopt transformer to model the subpart-part relationships, where the subpart capsules $\Theta^{s}_k:\{p_k^s, \theta_k^s, T_k^s, V_k^s, f_k^s\}$ are sent as input sequences to estimate part capsules $\Theta^{p}_m:\{T_m^p, V_m^p, r_m^p\}$:
	\begin{figure}
		\centering
		\includegraphics[width=1.0\linewidth]{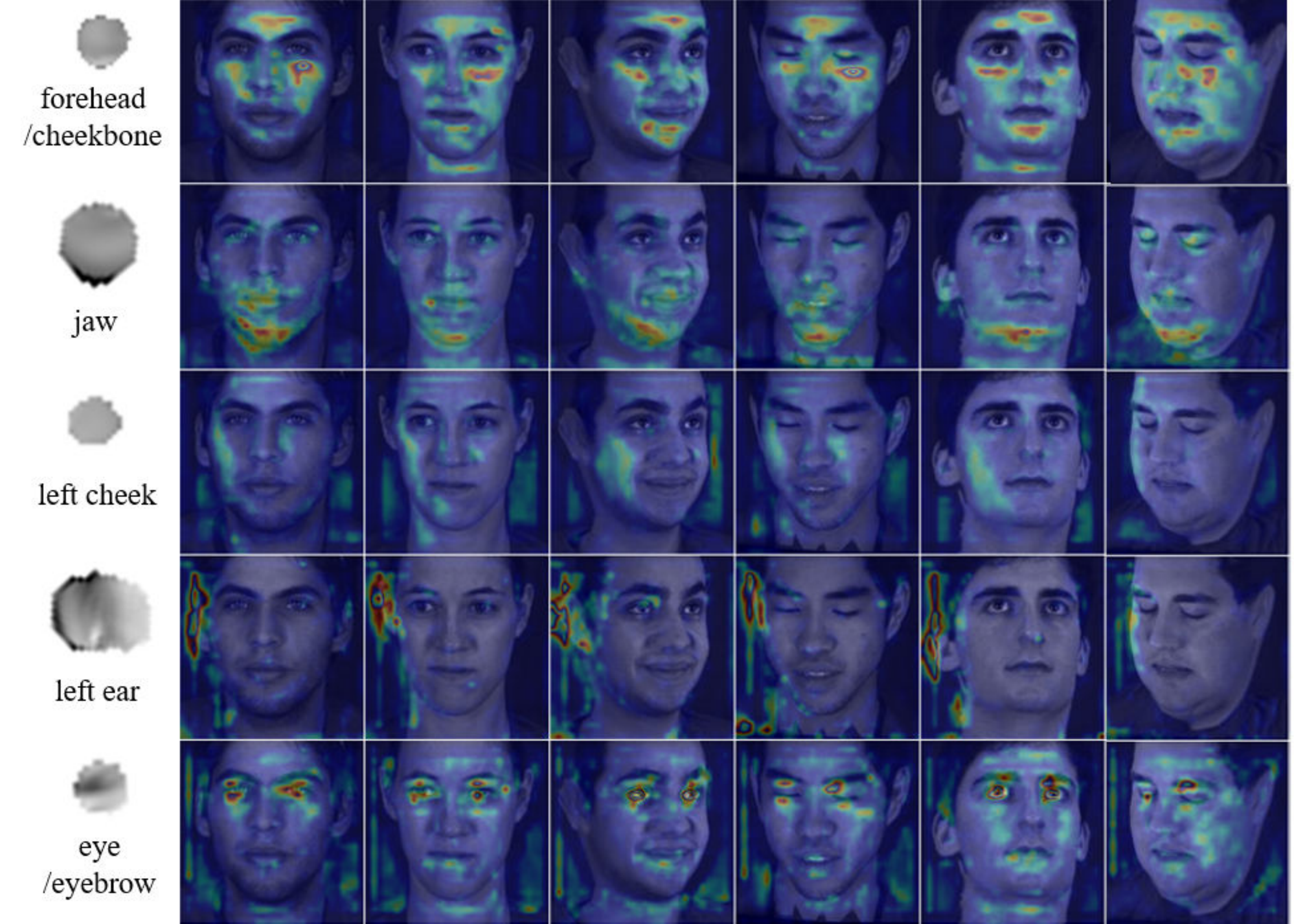}
		
		\caption{Examples of subparts and their corresponding heatmaps. Some smaller subparts lack semantics due to their ambiguity at different positions.}
		\label{fig-heatmap}
	\end{figure}
	\begin{equation}
		\begin{aligned}\label{eq-TPM}
			r^{p}_1, r^{p}_2, ..., r^{p}_M &= \rm{Transformer}(\Theta^{s}_1, \Theta^{s}_2, ..., \Theta^{s}_K),\\
			T^p_m &= \sum_{k}r^p_{m_k}p_k^sC^s_k\hat{T}_{k}^s,\\
			V^p_m &= \sum_{k}r^p_{m_k}p_k^s\hat{V}_{k}^s,
		\end{aligned}
	\end{equation}
	where $r_m^p=[r_{m_1}^p, ..., r_{m_K}^p]$ describes the probability that subpart capsules belongs to the $m$-th part. During training, we utilize K-means~\cite{pedregosa2011kmeans} to generate pseudo subpart-part relationships for the transformer and the $\mathcal{L}_{cls}$ can be formulated as:
	\begin{equation}
		\begin{aligned}\label{equ-cluster}
			\mathcal{L}_{cls} &= -\frac{1}{K}\sum_{k}\sum_{m}\check{r}_{m_k}^plog(r_{m_k}^p),\\
			\check{r}_m^p &=\mathrm{K\mbox{-}means}\left(\left[\hat{V}_{k}^s + t_{x_k}^s; \hat{V}_{k}^s + t_{y_k}^s\right]\right),
		\end{aligned}
	\end{equation}
	where $t_{x_k}^s, t_{y_k}^s$ are the translation parameters derived from the pose params $\theta_k^s$. Besides, we also assign a silhouette loss $\mathcal{L}_{silh}$ to constrain the shape consistency across samples:
	\begin{equation}\label{equ-silh}
		\mathcal{L}_{silh} = \frac{1}{M}\sum_{m}\Vert V_m^p - \check{V}_m^p \Vert_2,
	\end{equation}
	where $\check{V}_m^p$ is the averaged visibility map for the mini-batch.
	
	As for subpart assembling, since each subpart capsule should only belong to one part capsule, we pernalize the sparsity of compositional relations with $\mathcal{L}_{rela}$:
	\begin{equation}
		\label{equ-Lpre-p}
		\mathcal{L}_{rela} = \frac{1}{K}\sum_{k}\left[-\sum_{m}r_{m_k}^plog(r_{m_k}^p)\right].
	\end{equation}
	
	The proposed TPM incorporates both appearance and geometric clues for subparts assembling so that the generated parts have more prominent semantics. The overall loss function for hierachy construction can be be formulated as:
	\begin{equation}
		\label{equ-part}
		\mathcal{L}_{part} =\lambda_{cls}\mathcal{L}_{cls} +\lambda_{silh}\mathcal{L}_{silh} + \lambda_{rela}\mathcal{L}_{rela},
	\end{equation}
	where $\lambda_{cls},\lambda_{silh}$ and $\lambda_{rela}$ are the hyper-parameters to combine different loss functions.
	
	\section{Experiments}
	
	\subsection{Implementation Details}
	\label{subsec:details}
	HP-Capsule contains two sub-networks: the subpart capsule autoencoder with VAF and the Transformer-based Parsing Module. We replace the four convolutional layers in PCAE~\cite{kosiorek2019stacked} with two Residual Blocks~\cite{he2016deep} as the capsule autoencoder. For the transformer, we adopt the slot attention~\cite{slot2020} with learnable slots as encoder and use MLP as decoder, where the iteration of slot attention is set to 3. We extract the foreground mask by SSFNet~\cite{zhang2020semantic} to concentrate on the part discovery. For fair comparisons, we do the same operation for other unsupervised face segmentation methods that required foreground masks. 
	
	During training, the capsule autoencoder is first trained to converge and then refined together with TPM. For optimization, we use Adam optimizer~\cite{adam} with $10^{-4}$ learning rate. All experiments are implemented on Pytorch with a single NVIDIA Tesla M40 GPU.
	
	\subsection{Datasets}
	\label{subsec:datasets}
	\textbf{BP4D.}\indent We evaluate our model on the BP4D~\cite{zhang2014bp4d} from FG3D dataset~\cite{zhu2020beyond}. After 3D augmentation by ~\cite{zhu2020beyond}, it contains 23,359 images from 41 subjects with various expressions and poses. We randomly choose 90\% for training and 10\% for testing. 
	
	\textbf{Multi-PIE.}\indent The Multi-PIE dataset~\cite{gross2010multi} contains 337 subjects with various expressions. The images are captured under 15 different views spaced in $15^{\circ}$ intervals. In this paper, we choose the images from the middle 5 cameras with flash to avoid extreme poses, which contains 8,974 images. In the experiments, Multi-PIE is used to test the generalization ability. We split the dataset as 90\% for training and 10\% for testing, as the evaluation of unsupervised face segmentation also needs samples to train the fitting model.
	
	\subsection{Evaluation Metrics}
	\label{subsec:metric}
	
	The Normalized Concentrated Distance (NCD) is used in the ablation study to estimate the concentration of parts, which is formulated as:
	\begin{equation}\label{equ-NCD}
		\begin{aligned}
			&\mathrm{NCD} = \frac{1}{N}\sum_{n}\sum_{i,j}\Vert (i,j)-(c_i^n,c_j^n)\Vert_2 \cdot S_{n_{i,j}}/z_n, \\
			&(c_i^n, c_j^n)=\left(\sum_{i,j}i\cdot S_{n_{i,j}}/z_n, \sum_{i,j}j\cdot S_{n_{i,j}}/z_n\right),
		\end{aligned}
	\end{equation}
	where $S_n$ is the covered region of $n$-th part, $(c_i^n, c_j^n)$ is the centroid of the $n$-th part, $z_n = \sum_{i,j}S_{n_{i,j}}$ is for normalization.  
	
	The Normalized Mean Error (NME) of landmarks is used as an alternative way to evaluate the parsing quality, which is formulated as:
	\begin{equation}\label{equ-NME}
		\begin{aligned}
			\mathrm{NME} = \frac{1}{N}\sum_{n}\frac{\Vert v_n-v_n^* \Vert_2}{d},
		\end{aligned}
	\end{equation}
	where $N$ is the number of landmarks, $v_n$ is the landmark predicted from the segmentation results,  $v_n^*$ is the ground truth and $d$ is the inter-ocular distance. Following SCOPS~\cite{hung2019scops}, $\rm{NME_{L}}$ uses the centroids of segmentation maps as landmarks and converts them to the human-annotated landmarks by a linear mapping. During the test, we map the predicted 5 landmarks to 5 and 68 ground-truth landmarks separately. However, the centroids of the segmentation maps are too coarse to measure the parsing manners. In this paper, we propose $\rm{NME_{DL}}$ to evaluate the semantic consistency on a detailed level. $\rm{NME_{DL}}$ uses a very shallow network to predict the landmarks directly from the segmentation maps, which contains only one Residual Block and one linear layer.  
	
	\begin{figure}
		\centering
		\includegraphics[width=1.0\linewidth]{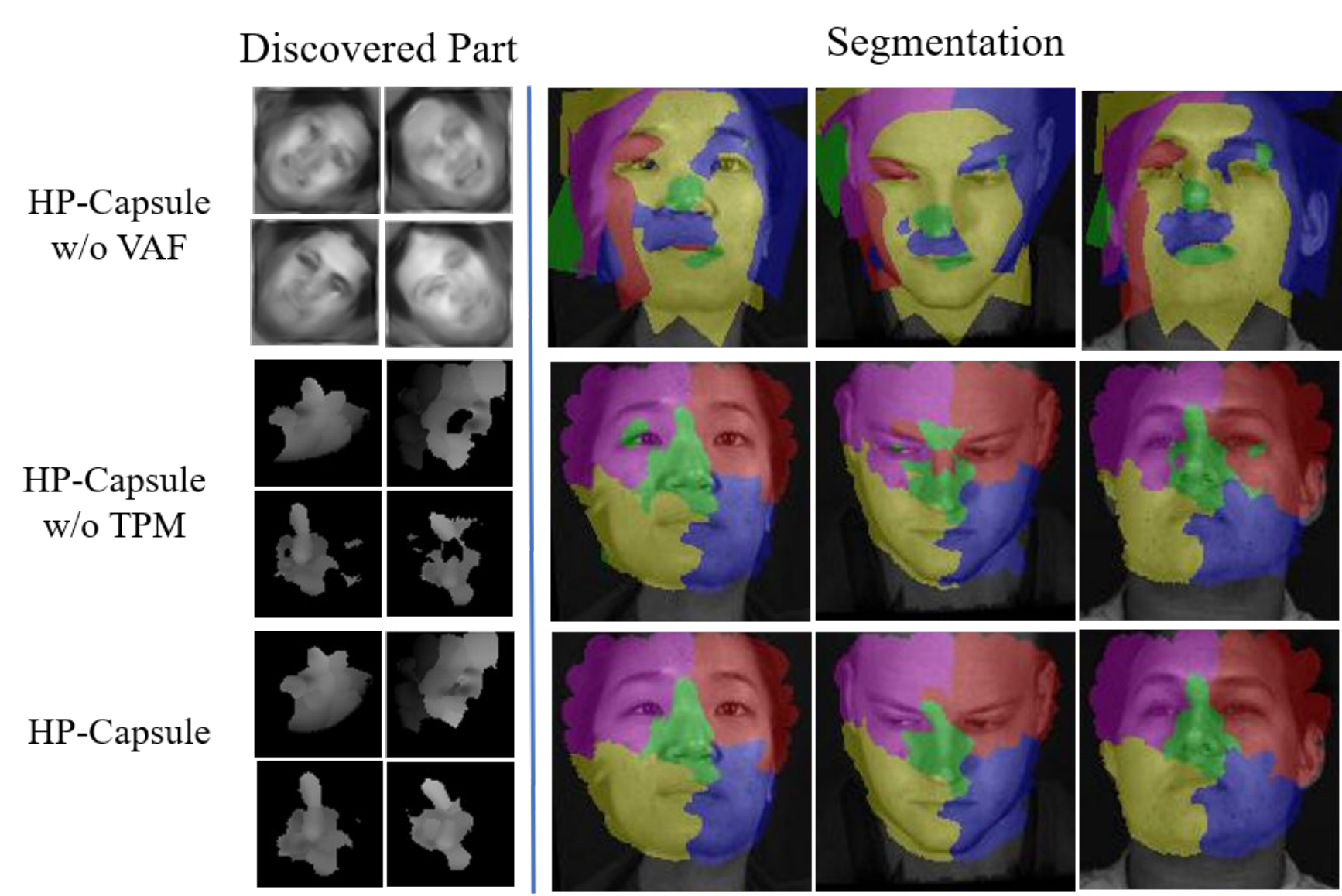}	
		\caption{The qualitative ablation study of HP-Capsule. We could see that VAF helps discover effective subparts and TPM improves the semantics of parts.}
		\label{fig-ablation}
	\end{figure}
	
	\begin{figure*}
		\centering
		\includegraphics[width=0.9\linewidth]{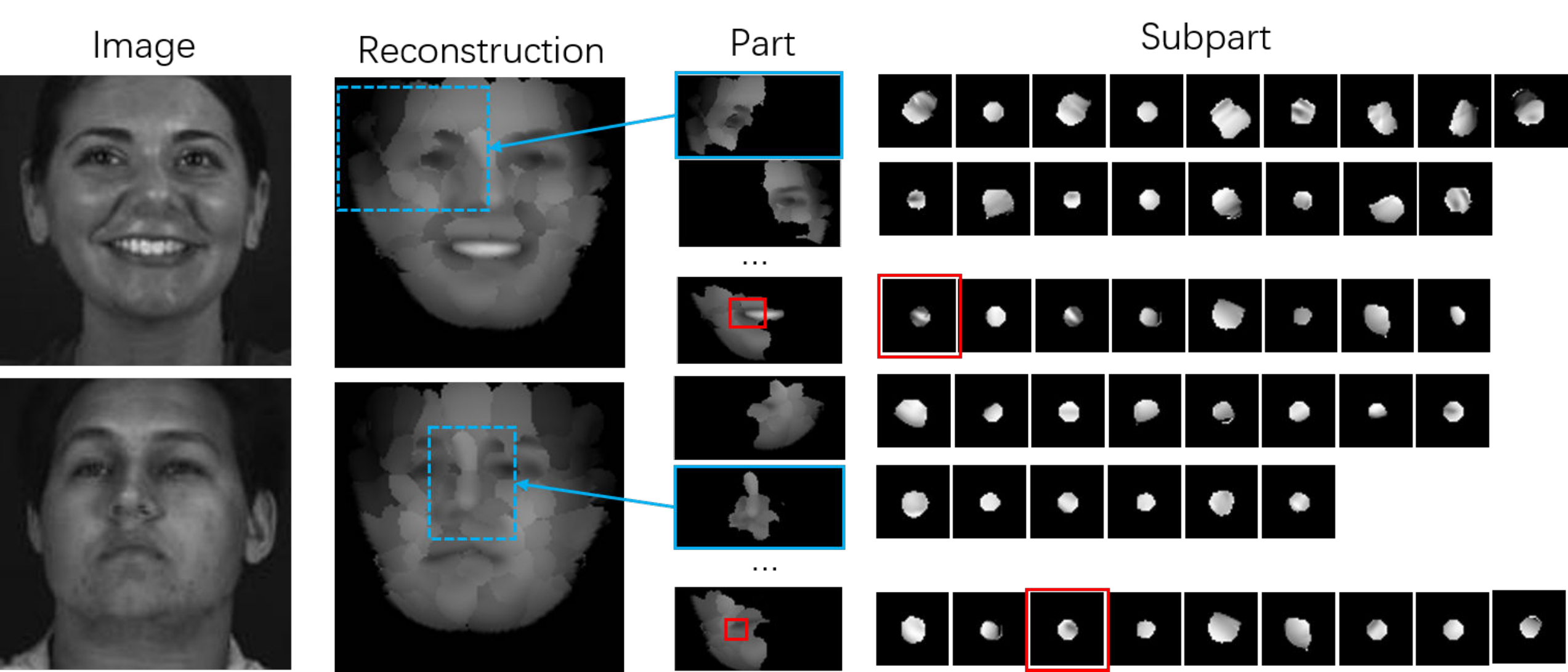}
		\caption{The hierarchical face parts discovered by HP-Capsule. For each input, HP-Capsule automatically selects a set of subparts to describe current object (marked with red rectangles) and aggregates them to get parts with more prominent semantics (marked with blue dot rectangles), constructing a bottom-up hierarchy. }
		\label{fig-discover}
	\end{figure*}
	
	\begin{table}
		\caption{The quantitative ablation study on BP4D. The results show the importance of VAF and TPM for hierarchical face part discovery.}
		\centering
		\begin{tabular}{ll|cc}
			\toprule
			VAF &  TPM & NCD &$\rm{NME_{DL}(\%)}$\\
			\midrule
			&   & 30.72  &- \\
			$\checkmark$  &   &19.27   &6.21 \\
			$\checkmark$  & $\checkmark$  &\textbf{18.79}  & \textbf{6.10}  \\
			\bottomrule
		\end{tabular}
		\label{tab-ablation}
	\end{table}

	\subsection{Ablation Study}
	\label{subsec:ablation}
	We perform the ablation study to show the importance of the VAF for subpart discovery and the TPM for hierarchy construction. 
	
	Figure~\ref{fig-ablation} shows the visualization results on face part discovery and segmentation. The first row shows that the original capsule network can not handle the faces, where the discovered parts are holistic and cover almost the whole face. In the second row, with the VAF, the network can capture effective local parts, as the regions with lower visibility are suppressed during training. However, the subparts have ambiguous semantics due to their limited expressive power, demonstrated in Figure~\ref{fig-heatmap}. Therefore, if only using the pseudo subpart-part relationships for hierarchy construction (Eqn.~\ref{equ-cluster}), the aggregated parts will suffer serious flaws and become ambiguous in semantics, shown in the second row. After introducing the TPM for refinement, the semantics of parts is improved. The importance of VAF and TPM is further validated by the quantitative evaluation in Table~\ref{tab-ablation}.

	\begin{figure}
	\centering
	\includegraphics[width=1.0\linewidth]{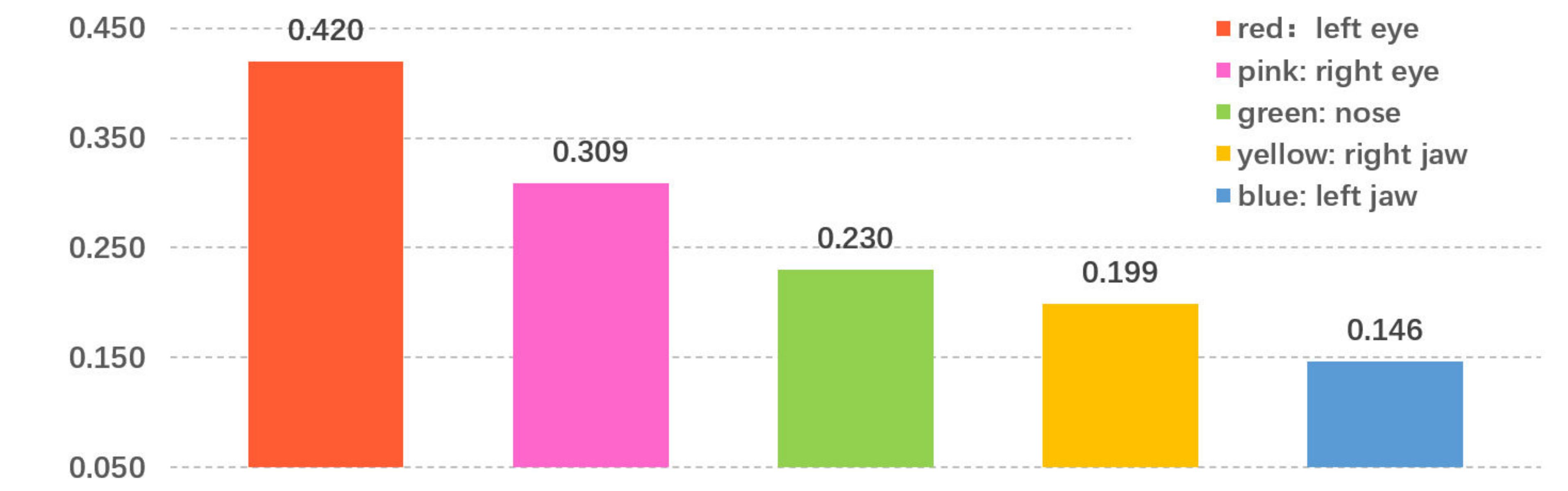}
	
	\caption{The importance of each region for face recognition. The color of columns is identical to segmentation maps. The regions with eyes are more important for recognition.}
	\label{fig-recog}
	\end{figure}	
	
	\subsection{Analysis on the Exploited Face Hierarchy}
	\label{subsec:exp-discovery}
	For each input, HP-Capsule automatically activates a subset of subpart capsules and aggregates them to higher-level part capsules, constructing a bottom-up hierarchy. 
	Figure~\ref{fig-discover} shows the hierarchical face parts discovered by HP-Capsule. It can be seen that HP-Capsule almost reconstructs the original images and parses them into five parts: nose, left side of mouth and cheek, right side of mouth and cheek, the left upper face containing left eye and left ear, and the right upper face. 
	The last column shows the corresponding subparts of each part. Through the templates defined in capsules, the discovered concepts can be directly visualized without any decoding operations, which makes them easier for us to understand.
	
	We also design a toy experiment to show which part capsule is important for face recognition.
	We assign each subpart capsule a trainable non-negative scalar $w_k$ as attention weight and send the weighted subpart capsule parameters $w_k\Theta_k^s$ to a linear classifier for face recognition: $y = \mathrm{Linear}(w_1\Theta_1^s, ..., w_K\Theta_K^s)$. The classifier is trained with softmax and L1 penalization for the sparsity of $w_k$. The averaged weights for each part are shown in Figure~\ref{fig-recog}. It can be seen that eye regions are most important for face recognition, which is consistent with the work of Williford \etal.~\cite{williford2020explainable} that shows eyes and nose contain more discriminative features for face recognition.

	\subsection{Comparison on Unsupervised Face Segmentation}
	\label{subsec:exp-segmentation}
	The covered regions of parts discovered by HP-Capsule can be used for the unsupervised face segmentation task, enabling the comparison with other state-of-the-art methods. 
	
	\textbf{Methods.}\indent The unsupervised face segmentation from unlabeled images is a challenging task that has not been well explored. DFF~\cite{collins2018deep} proposes to use non-negative matrix factorization upon the CNN features to discover the semantic concepts, which needs to optimize on the whole datasets during inference to keep semantic consistency. SCOPS~\cite{hung2019scops} incorporates the invariance between TPS transformation as clues and proposes a framework with several loss functions for unsupervised segmentation. However, their method mainly relies on the concentration loss, which tends to cut the images into the same region regardless of the face poses.    

	\begin{figure}
	\centering
	\includegraphics[width=1.0\linewidth]{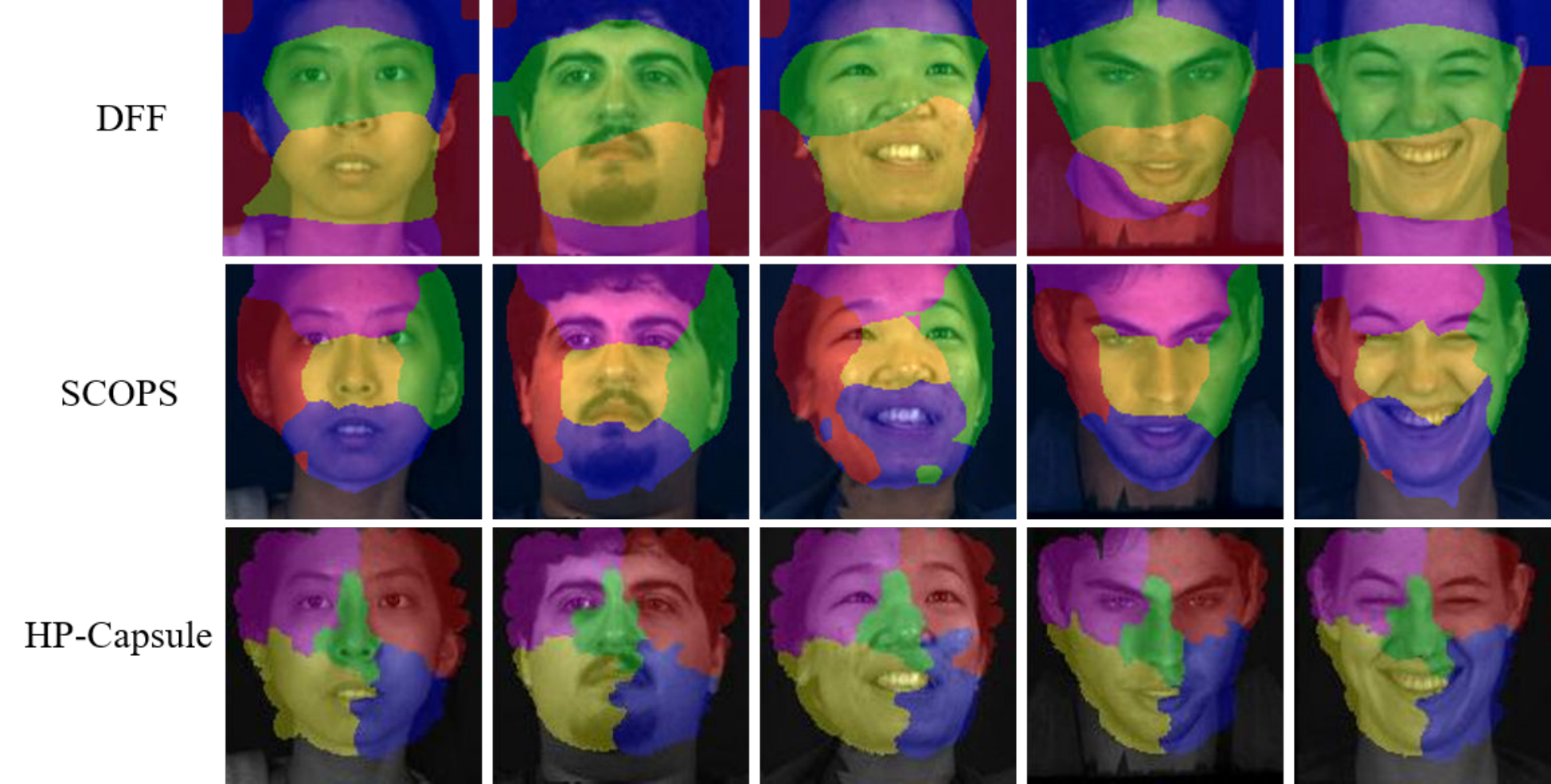}
	
	\caption{The qualitative comparsion of unsupervised face segmentation on BP4D. HP-Capsule shows better semantic consistency across samples.}
	\label{fig-BP4D}
	\end{figure}

	\begin{figure}
		\centering
		\includegraphics[width=1.0\linewidth]{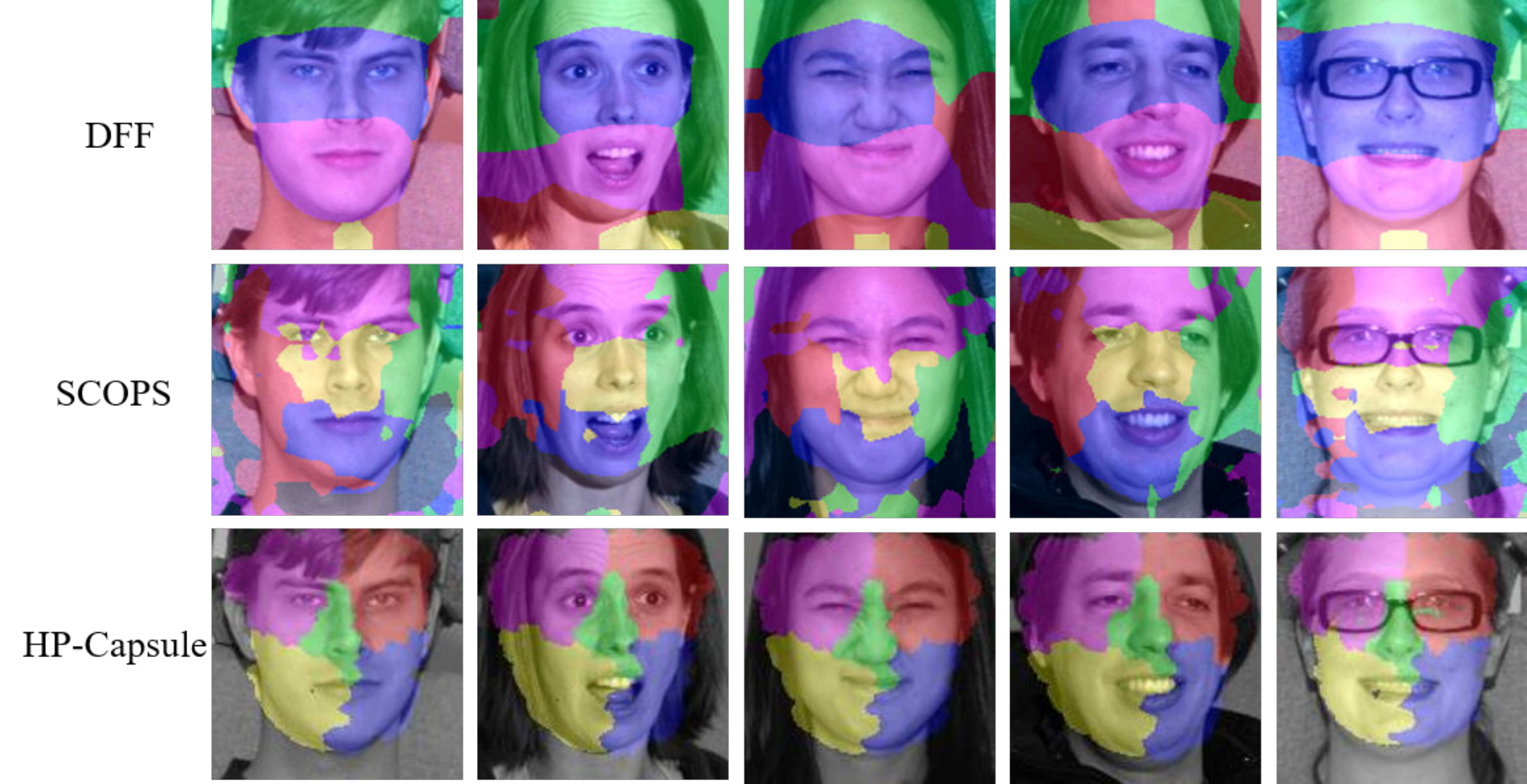}
		
		\caption{The qualitative comparison of unsupervised face segmentation on Multi-PIE. HP-Capsule shows better stability when generalized to another dataset. }
		\label{fig-MultiPIE}
	\end{figure}

	\begin{table}
		\caption{The quantitative comparison  of unsupervised face segmentation on BP4D. $\rm{NME_{L}}$(\%) and $\rm{NME_{DL}}$(\%) evaluate the semantic consistency of landmarks.}
		\centering
			\begin{tabular}{cccc}
				\toprule
				\multirow{2}{*}{Method}  & \multicolumn{2}{c}{$\rm{NME_{L}}$}  & \multirow{2}{*}{$\rm{NME_{DL}}$}\\
				\cmidrule(lr){2-3}   &$\mathrm{N_{GT}=5}$ & $\mathrm{N_{GT}=68}$ &  \\
				\midrule
				DFF ~\cite{collins2018deep} & 18.85  &18.62    & 12.26  \\
				SCOPS ~\cite{hung2019scops}  & 9.10  & 9.67  &  6.74 \\
				HP-Capsule &\textbf{8.81}  &\textbf{9.11}     & \textbf{6.10}   \\
				\bottomrule
			\end{tabular}
		\label{tab-BP4D}
	\end{table}
	
	\textbf{Evaluation on BP4D.}\indent From Figure~\ref{fig-BP4D}, it can be seen that without any spatial transformation as clues, HP-Capsule still outperforms other methods with better semantic consistency. As shown in the second row, the results of SCOPS sometimes take the right eye as the right green part, sometimes take it as the top pink part or even split it from the middle. In comparison, HP-Capsule keeps the same parsing manners under different poses. This is because the shared templates of HP-Capsule encode the frequently observed patterns and the intrinsic geometric clues are captured for better image reconstruction. 
	Results in Table~\ref{tab-BP4D} also validate the effectiveness of our method.
	
	\textbf{Evaluation on Multi-PIE.}\indent To further investigate the generalization of methods, we employ Multi-PIE for cross-data evaluation. In Figure~\ref{fig-MultiPIE}, we can see that the SCOPS cannot separate the foreground well even it has already been trained with the ground-truth masks of BP4D. Different from previous methods, HP-Capsule learns to estimate face parts with explicit semantics and generalizes well on the Multi-PIE dataset, which is also validated in Table~\ref{tab-MultiPIE}.

	\begin{table}
		\caption{The quantitative comparison  of unsupervised face segmentation on Multi-PIE. $\rm{NME_{L}}$(\%) and $\rm{NME_{DL}}$(\%) evaluate the semantic consistency of landmarks.}
		\centering
			\begin{tabular}{cccc}
				\toprule
				\multirow{2}{*}{Method}  & \multicolumn{2}{c}{$\rm{NME_{L}}$} &\multirow{2}{*}{$\rm{NME_{DL}}$}\\
				\cmidrule(lr){2-3}   &$\mathrm{N_{GT}=5}$ & $\mathrm{N_{GT}=68}$    &  \\
				\midrule
				DFF ~\cite{collins2018deep}   & 21.88  & 20.21  & 17.35  \\
				SCOPS ~\cite{hung2019scops}  &16.15  &15.25  & 13.54   \\
				HP-Capsule &\textbf{12.34} &\textbf{12.30}    & \textbf{11.57}  \\
				\bottomrule
			\end{tabular}
		\label{tab-MultiPIE}
	\end{table}

	\section{Discussion and Conclusion}
	\label{sec:conclusion}
	In this paper, we propose the HP-Capsule for unsupervised face part discovery. By incorporating the Visibility Activation Function and the Transformer-based Parsing Module, the network successfully discovers face parts and constructs the face hierarchy from the unlabeled images. This work extends the application of capsule networks from digits to human faces and provides an insight into the visual perception mechanism of networks. Besides, HP-Capsule gives unsupervised face segmentation results by the covered regions of part capsules, enabling the comparison with other state-of-the-art methods. Experiments on BP4D and Multi-PIE show the superior performance of our method.
	
	\textbf{Limitation.}\indent HP-Capsule presents the face hierarchy with a set of visualizable templates, whose representation power is limited by the number of the templates. Improving the representation ability of templates while maintaining their interpretability remains a challenging task and deserves further studies in future work.
		
	\textbf{Broader Impacts.}\indent The proposed method is a generative model that discovers face hierarchy based on the learned statistics of the training dataset, which will reflect biases in those data, including ones with negative societal impacts. These issues are worth further consideration if the capsules are utilized to generate images.
	
	\vspace{8px}
	
	\noindent\textbf{Acknowledgement.} This work was supported in part by the National Key Research \& Development Program (No. 2020YFC2003901), Chinese National Natural Science Foundation Projects \#62176256, \#61876178, \#61976229, \#62106264, the Youth Innovation Promotion Association CAS (\#Y2021131) and the InnoHK program.

	\clearpage
	{\small
		\bibliographystyle{ieee_fullname}
		\bibliography{egbib}
	}

\clearpage
\newpage
\appendix

\section{More Implementation Details}
\label{sec:details}
\subsection{Training Details}
We set the number of subpart capsules $K=75$, the number of part capsules $M=5$, $\gamma=0.5$ for VAF, $\tau=16$ for subpart presence sparsity loss. $\lambda_{cen}, \lambda_{cls},\lambda_{silh}$ for loss combination are set to $0.5, 10^2, 10^3$, and other hyper-parameters are set to 1. The input images are resized to $128\times 128$. The subpart templates are set to $40\times 40$ and the part templates are set to $128\times 128$.

\subsection{Details of Template Transformation}
During the subpart discovery, our Hierarchical Parsing Capsule Network (HP-Capsule) performs affine transformation with pose $\theta^s$ to transform each subpart template into the image space. In our implementation, $\theta^s=(s^s, h^s, a_x^s, a_y^s, t_x^s, t_y^s)$ is a 6-tuple, including $s^s$ for scaling, $h^s$ for shearing, $(a_x^s,a_y^s)$ for rotation, and $(t_x^s,t_x^s)$ for translation. The transformation matrix can be formulated as:
\begin{equation}
	\label{equ-affine}
	A = 
	\begin{bmatrix}
		s^scosa & -s^ssina+s^sh^scosa & t_x^s\\
		s^ssina & s^scosa+s^sh^ssina & t_y^s\\
		0 & 0 & 1 	
	\end{bmatrix}
\end{equation}
where $(cosa, sina) = (a_x^s, a_y^s)/\Vert(a_x^s, a_y^s) \Vert_2$. We use $a_x^s$ and $a_y^s$ to estimate the rotation angle to avoid the continuity issue~\cite{zhou2019continuity,Gao2021ucos}.

\subsection{More Details about $\rm{NME_{DL}}$}
We propose a new evaluation metric $\rm{NME_{DL}}$ in the main paper to evaluate the unsupervised segmentation results on a detailed level, which uses a very shallow network to directly predict landmarks from the segmentation maps. The shallow network is trained using Adam with $10^{-4}$ learning rate for 50 epochs and all the input segmentation maps are resized to $32\times 32$. During the evaluation, we choose three different shallow networks for $\rm{NME_{DL}}$, including two convolution layers, one Residual Block~\cite{he2016deep}, and two Residual Blocks. Each of them is followed by a linear layer. Table~\ref{tab-NMEDL} shows the sophisticated evaluation by $\rm{NME_{DL}}$ with different architectures on BP4D. It can be seen that our method surpasses other methods with better semantic consistency.

\begin{table}
	\caption{The quantitative comparison  of unsupervised face segmentation on BP4D. $\rm{NME_{DL}}$(\%) is implemented with different architectures to evaluate the semantic consistency of landmarks.}
	\centering
	\begin{tabular}{cccc}
		\toprule
		Method  & 2 Conv & 1 ResBlock & 2 ResBlocks \\
		\midrule
		DFF ~\cite{collins2018deep} & 14.53  &12.26    &12.28   \\
		SCOPS ~\cite{hung2019scops}  &7.91   & 6.74  & 6.94  \\
		HP-Capsule &\textbf{6.83}  &\textbf{6.10}     & \textbf{6.26}   \\
		\bottomrule
	\end{tabular}
	\label{tab-NMEDL}
\end{table}

\section{More Analysis on the Exploited Face Hierarchy}
\begin{figure}
	\centering
	\includegraphics[width=0.9\linewidth]{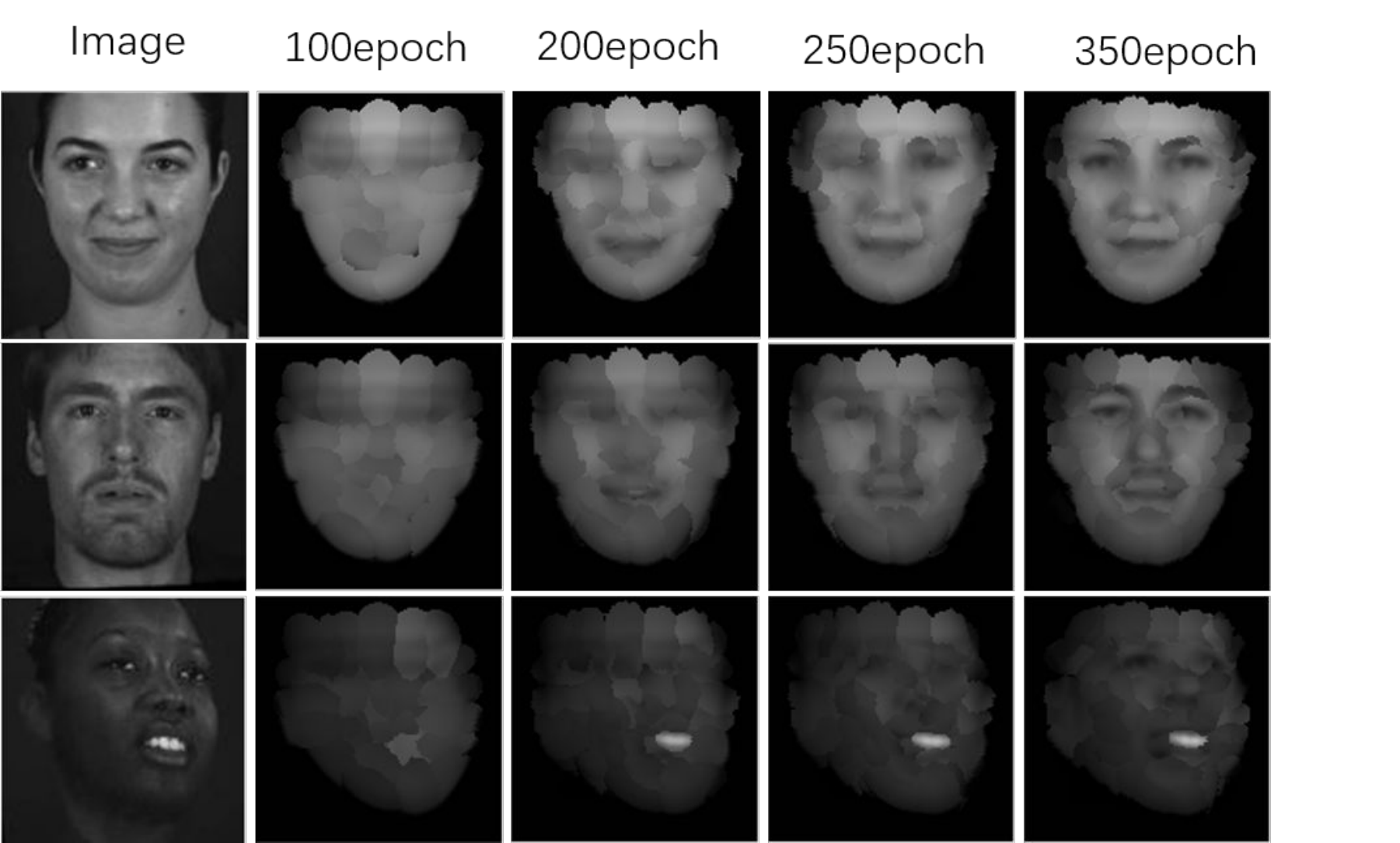}
	
	\caption{The learning procedure of HP-Capsule. The network captures the facial features in the sequence of face contour, mouth, nose, and finally eyes, which indicates the nose and eyes might contain more identity-specific information.}
	\label{fig-rec}
\end{figure}

To further explore the visual perception mechanism of neural networks, we visualize the learning procedure of HP-Capsule in Figure~\ref{fig-rec}. It can be seen that the network captures the facial features in the sequences of face contour, mouth, nose, and finally eyes. One reasonable guess is that nose and eyes contain more identity-related information, making them more difficult to be reconstructed. This conclusion is also consistent with the work of Williford \etal.~\cite{williford2020explainable} that shows nose and eyes contain more discriminative features for face recognition. 

\begin{figure}
	\centering
	\includegraphics[width=0.9\linewidth]{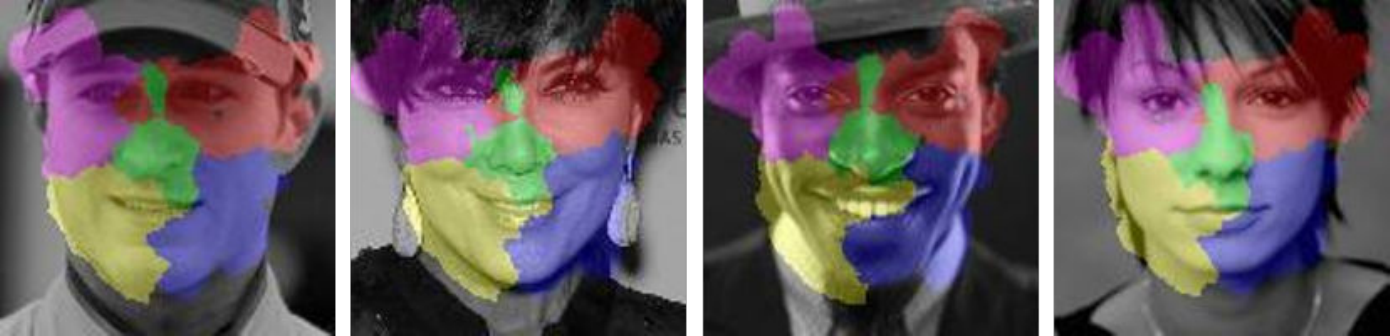}	
	\caption{The segmentation results of HP-Capsule on CelebA.}
	\label{fig-seg-CelebA}
\end{figure}

\section{More Visualization Results}
\label{sec:visual}
To validate the potential of HP-Capsule under the in-the-wild scenarios, we evaluate our method on CelebA\cite{liu2015deep}.
As shown in Figure~\ref{fig-seg-CelebA}, our method can also keep semantic consistency among the in-the-wild images.

We provide more visualization results of our HP-Capsule on BP4D~\cite{zhang2014bp4d,zhu2020beyond} and Multi-PIE~\cite{gross2010multi}. Figure~\ref{fig-parse-BP4D} shows the hierarchical face parts discovered by HP-Capsule. Figure~\ref{fig-seg-BP4D} and Figure~\ref{fig-seg-MultipiE} show the unsupervised segmentation results on BP4D and Multi-PIE. It can be seen that our method can discover the face hierarchy directly from the unlabeled images and keep semantic consistency across different samples.

\begin{figure}
	\centering
	\includegraphics[width=0.9\linewidth]{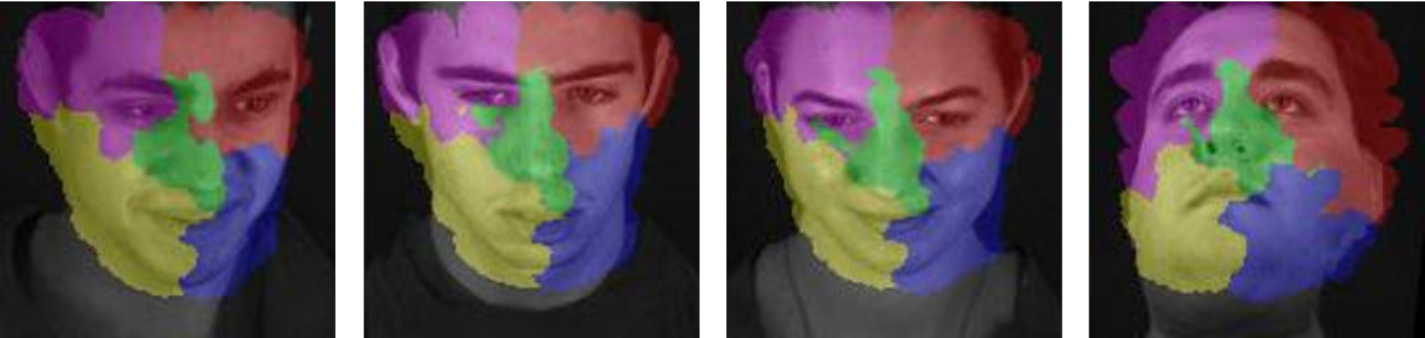}	
	\caption{Some failed results of unsupervised face segmentation on BP4D. The parts of faces in large poses may still have flaws after refinement.}
	\label{fig-seg-BP4D-flaw}
\end{figure}

\begin{figure*}
	\centering
	\includegraphics[width=1.05\linewidth]{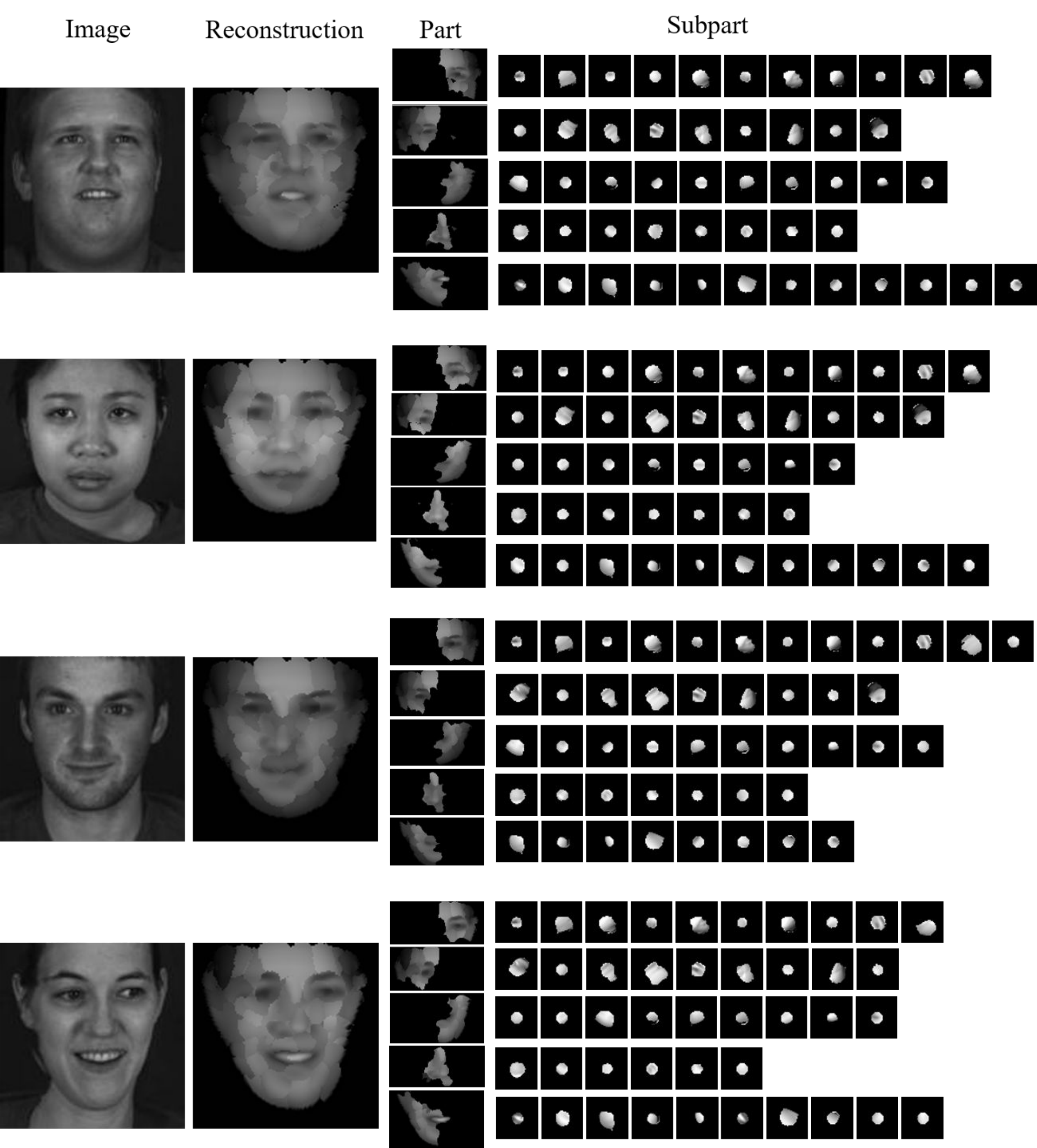}	
	\caption{The hierarchical face parts discovered by HP-Capsule. For each input, HP-Capsule automatically selects a set of subparts to describe the current object and aggregates them to get parts with more prominent semantics.}
	\label{fig-parse-BP4D}
\end{figure*}

\begin{figure*}
	\centering
	\includegraphics[width=0.95\linewidth]{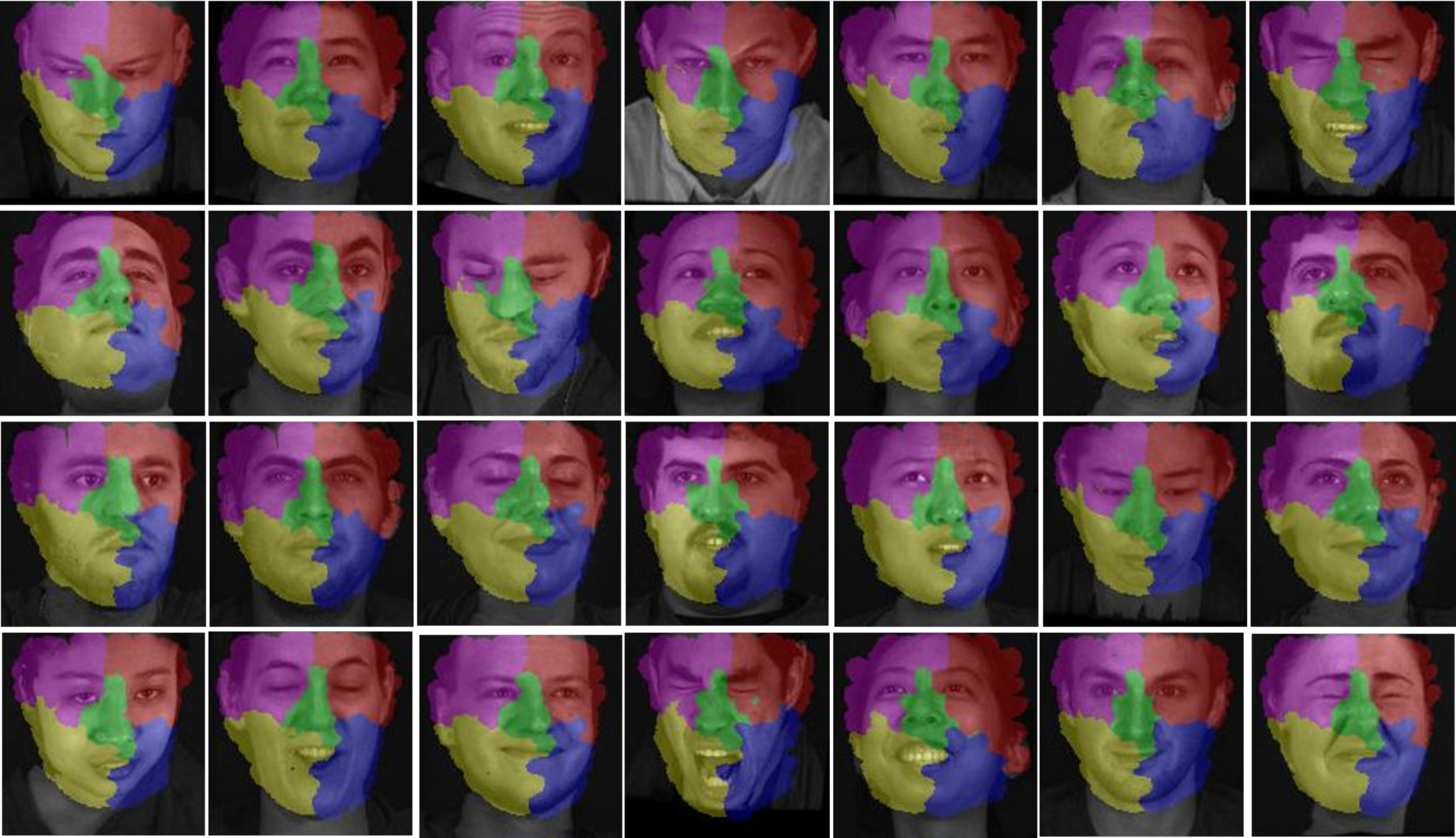}	
	\caption{Visualization results of unsupervised face segmentation on BP4D.}
	\label{fig-seg-BP4D}
	\vspace{10px}
\end{figure*}

\begin{figure*}
	\centering
	\includegraphics[width=0.95\linewidth]{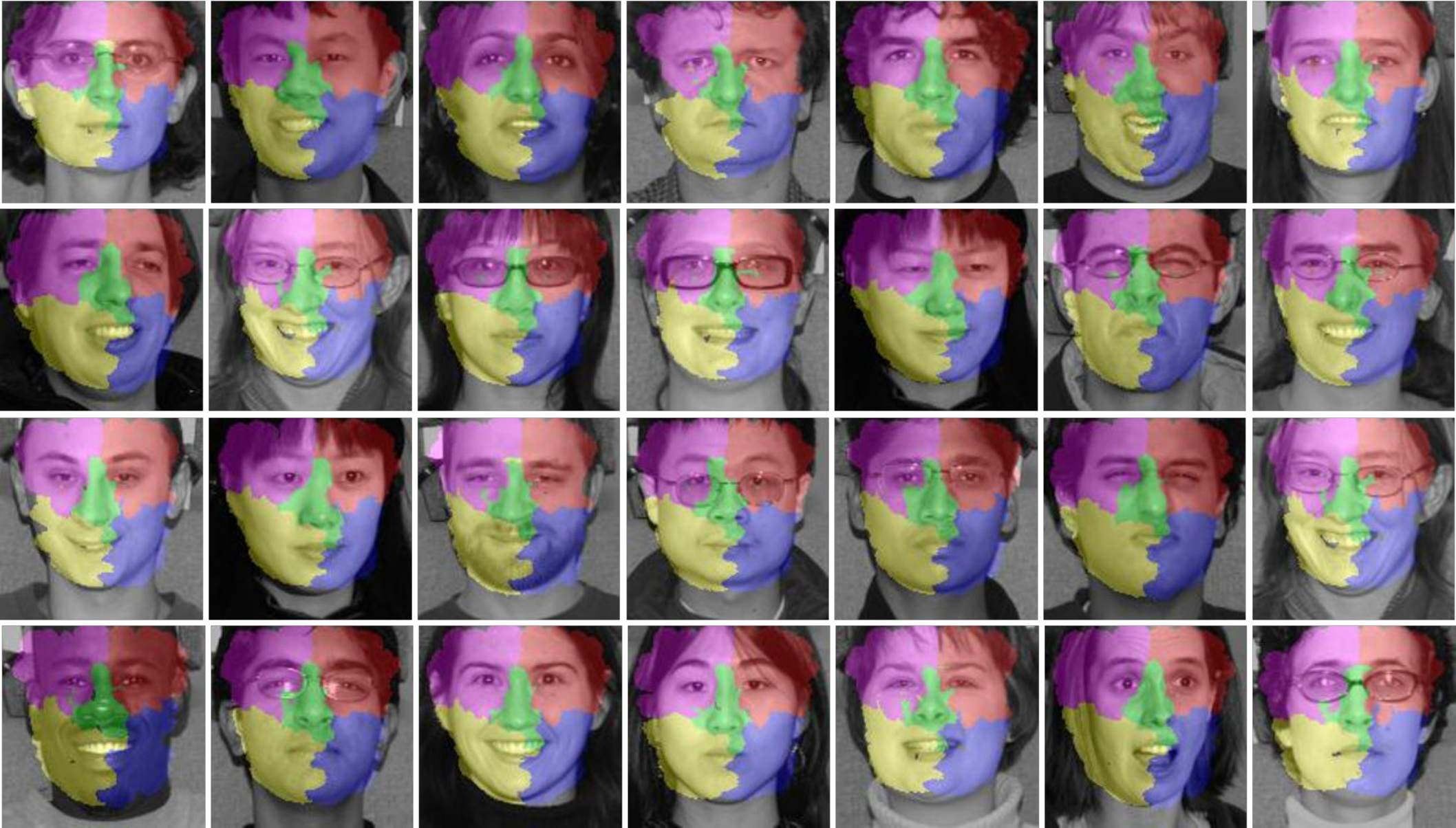}	
	\caption{Visualization results of unsupervised face segmentation on Multi-PIE.}
	\label{fig-seg-MultipiE}
	\vspace{10px}
\end{figure*}

We also show some typical failure cases in Figure~\ref{fig-seg-BP4D-flaw}. Although the semantics of parts have been improved after introducing the Transformer-based Parsing Module for refinement, there may still be flaws on some faces in large poses.

\vspace{2px}

\end{document}